\documentclass{article}
\usepackage{spconf,amsmath,graphicx}
\usepackage{xcolor}
\usepackage{hyperref}
\usepackage[nocompress]{cite}

\title{Classification of Spontaneous and Scripted Speech\\ for Multilingual Audio}
\name{Shahar Elisha\textsuperscript{1,2},  Andrew McDowell\textsuperscript{1}, Mariano Beguerisse-D\'iaz\textsuperscript{1}, Emmanouil Benetos\textsuperscript{2}\thanks{We thank K. Lingley for his support of this project. We thank J. Correia, R. Dall, D. Hristova, R. Jones, J. Karlgren, D. Korkinof, A. Lima, L. Lin, S. Reddy, B. Regan, F. Vaughn, and Y. Yu for their useful feedback.}}

\address{\textsuperscript{1}Spotify Ltd \\
\textsuperscript{2}Centre for Digital Music, Queen Mary University of London}

\begin{document}
\ninept
\maketitle
\begin{abstract}
Distinguishing scripted from spontaneous speech is an essential tool for better understanding how speech styles influence speech processing research. It can also improve recommendation systems and discovery experiences for media users through better segmentation of large recorded speech catalogues. This paper addresses the challenge of building a classifier that generalises well across different formats and languages. We systematically evaluate models ranging from traditional, handcrafted acoustic and prosodic features to advanced audio transformers, utilising a large, multilingual proprietary podcast dataset for training and validation. We break down the performance of each model across 11 language groups to evaluate cross-lingual biases. Our experimental analysis extends to publicly available datasets to assess the models' generalisability to non-podcast domains. Our results indicate that transformer-based models consistently outperform traditional feature-based techniques, achieving state-of-the-art performance in distinguishing between scripted and spontaneous speech across various languages.

\end{abstract}
\section{Introduction}
\label{sec:intro}

Large scale audio-based classification of multilingual speech is a technical challenge that can enable both academic research and industrial applications. For example, distinguishing between speech patterns, styles and tones can underpin contextual recommendations, advanced search and enable assistive technologies, to name a few. In this paper, we develop and evaluate models to distinguish between scripted and spontaneous speech on a large, multilingual podcast dataset. While the differences between read and improvised speech have been investigated in the linguistics literature~\cite{spontaneous_or_read, LAAN199743, f0_speaking_style, Tucker_Mukai_2023}, we approach the problem from a computational perspective, with an emphasis on large-scale, diverse data, and modern audio models.
The motivation for distinguishing between spontaneous and scripted speech in multiple languages is manifold: increasingly large audio repositories require efficient, performant, and generalisable content classification models. We also want to address the need to understand how speaking styles and languages affect various speech models. 

In recent years, media services have offered an increasingly large catalogue of recorded speech in a range of formats, such as podcasts, audiobooks, lectures and archival collections of broadcasts. As user preferences range over stylistic variations~\cite{martikainen2022exploring}, enriching catalogues with style and other labels (e.g., as in video~\cite{deldjoo2016content, Deldjoo2020} and visual arts~\cite{xie2022artistic} applications) results in better recommendation systems and an improved discovery experience for listeners. Building a high-performant and efficient scripted/spontaneous speech classifier is an initial and important step in this direction. In addition, some catalogues, such as those in major streaming platforms, encompass content from around the world, covering a wide range of languages, cultures, and formats. Therefore, we are particularly interested in developing methods that perform well across languages, topics and formats, as well as understanding the biases and shortcomings of the current state-of-the-art.

More so, there is an increasing body of work suggesting a direct impact of speech style on speech processing research~\cite{WAGNER20151}, influencing tasks such as speaker identity~\cite{speech_variation_speaker_rec_implications}, emotion recognition~\cite{audeering2021multimodal}, disease recognition~\cite{luz20_interspeech}, and speech generation~\cite{williams19c_interspeech}. For example, one study reported that anger detection performs better on acted speech compared to natural conversations~\cite{audeering2021multimodal}. This raises a question of how we bias speech models with curated datasets, and calls for a deeper understanding of how speech styles interplay with various speech models. We believe that the ability to reliably distinguish between scripted and spontaneous speech is needed to evaluate the effects of speaking style on downstream speech processing models. The implementation can encourage speech researchers to measure and mitigate biases, and thereby build more generalisable models. This is particularly important today given the explosion in popularity, accessibility and performance of transformer models, the increasing availability of large datasets of natural and diverse speech, and the presence and function of speech research in several aspects of day-to-day life~\cite{Tucker_Mukai_2023, ethical_awareness_paralinguistics}. 

Our contributions can be divided into three main parts: first, we train and compare the performance of different audio models for spontaneous vs scripted speech classification on a large, proprietary, multilingual podcast dataset. Second, we break down the performance of the models by language to assess multilinguistic biases. Third, we measure the models' abilities to generalise to non-podcast domains using publicly available datasets. We provide a comparison of handcrafted acoustic and prosodic features to modern speech representation models, and an evaluation of how our results generalise to a range of data sources and languages. We find that transformer-based models achieve state-of-the-art performance in distinguishing scripted from spontaneous speech across different domains and languages. To our knowledge, this is the first effort to systematically study state-of-the-art models to distinguish spontaneous from scripted speech, with a novel investigation into the nuances of different languages through a large and diverse, multilingual dataset.

The rest of the paper is structured in the following way: we review related works in Section~\ref{sec:related_works}, outline details of our experimental setup in Section~\ref{sec:experimental_setup}, provide a breakdown of the experimental results in Section~\ref{sec:results}, and conclude with a discussion on our contributions and future works in Section~\ref{sec:conclusion}.

\section{Related works}
\label{sec:related_works}

The relationship between various acoustic and prosodic features and distinct speaking styles, in particular read versus spontaneous speech, has been extensively investigated over the last few decades~\cite{spontaneous_or_read, LAAN199743, f0_speaking_style, Tucker_Mukai_2023}. While no single feature holds substantial predictive power, prosodic features such as pitch (F0) contours and articulation rates seem indicative of spontaneous speech~\cite{spontaneous_or_read, f0_speaking_style, speakerStyle_Laan_1997}. The salience and variability of these features can differ between languages, speaking situations, environments, and even individual speakers~\cite{Tucker_Mukai_2023, speech_variation_speaker_rec_implications}.  

More recently, researchers have leveraged automated
tools to understand the differences between speaking styles on larger datasets. A couple of studies explored building a multidimensional space of acoustic and prosodic features to characterise a range of speaking styles (e.g. conversations, audiobooks, speeches)~\cite{ryant16_interspeech, christodoulides20_speechprosody}, while another study trained a binary classifier to distinguish read and spontaneous speech~\cite{speaking_style_classification}. They all extracted a range of handcrafted features such as pitch ranges, speaker rates, duration of speech and silence, amongst other temporal, pitch, and conversational dynamics measures. It is generally agreed that the distributions of speech and silence segments are salient features to discriminate spontaneous from narrated speech. Other salient features, according to Christodoulides~\cite{christodoulides20_speechprosody}, relate to interactivity (speaker turns and overlapping speech), pitch ranges and contours (e.g. rises and falls), and other temporal characteristics (e.g. speaker rate, articulation ratio), however Ryant and Liberman~\cite{ryant16_interspeech} find that pitch range is less relevant. Spontaneous conversations, compared to other speaking styles (e.g. political and professional public speeches), are found to be harder to classify due to larger speaker-specific variations~\cite{christodoulides20_speechprosody, speech_variation_speaker_rec_implications, speaking_style_classification}. More concretely, Veiga et al.~\cite{speaking_style_classification} reported accuracies of 93.7\% and 69.5\% on read and spontaneous speech segments, respectively. 

The studies drew similar conclusions based on datasets in different languages (Portuguese; French; and English, Chinese, and Spanish). Ryant and Liberman's was the only multilingual study; they report that the acoustic patterns are generally consistent across the three languages, but noted differences in the duration of silences between Spanish and English audiobooks. This requires further investigation due to the small sample size of Spanish speakers ($n=6$), and further encourages the need to systematically investigate cross-lingual differences on a larger scale, with a wider range of languages.

Finally, all the experiments above explore handcrafted acoustic and prosodic features to assess the differences between speaking styles. To our knowledge, modern audio embedding models have yet to be applied to speech style classification. While they offer reduced interpretability, they often allow efficient, larger-scale studies, and show state-of-the-art performance on a range of adjacent tasks~\cite{ethical_awareness_paralinguistics}. For example, speech emotion recognition has an endless collection of studies experimenting with a range of audio features; while handcrafted feature sets (e.g. eGeMAPSv2~\cite{opensmile_gemaps}) were the common approach to the task~\cite{schuller2018speech}, the emergence of large transformer models (e.g. Whisper~\cite{whisper} and Wav2Vec 2.0~\cite{wav2vec2}) attain much better performance~\cite{latif2021sersurvey, whisper_ser}. Genre classification using YAMNet features \cite{howard2017mobilenets} has been reportedly successful for distinguishing between different television programmes (e.g. comedy, drama, news, sports, weather, children's) with accuracy as high as 93.7\%~\cite{bbc_program}. Whisper, while developed as an Automatic Speech Recognition (ASR) model, has been effectively repurposed as a classification model, and shows state-of-the-art performance on tasks such as disfluency identification and stuttered speech classification \cite{ameer2023whisper}. This study reports superior performance using Whisper compared to Wav2Vec 2.0, comparable to similar studies on different tasks, such as DeepFake detection \cite{deepfake_detection_whisper}, vocal intensity categorisation \cite{kodali23_interspeech}, and emotional nonverbal vocalisation detection \cite{nonverbal_vocalisation}.  

\section{Experimental setup}
\label{sec:experimental_setup}

\subsection{Data}
\label{sec:data}

\begin{figure}[t]
  \centering
  \includegraphics[width=\linewidth]{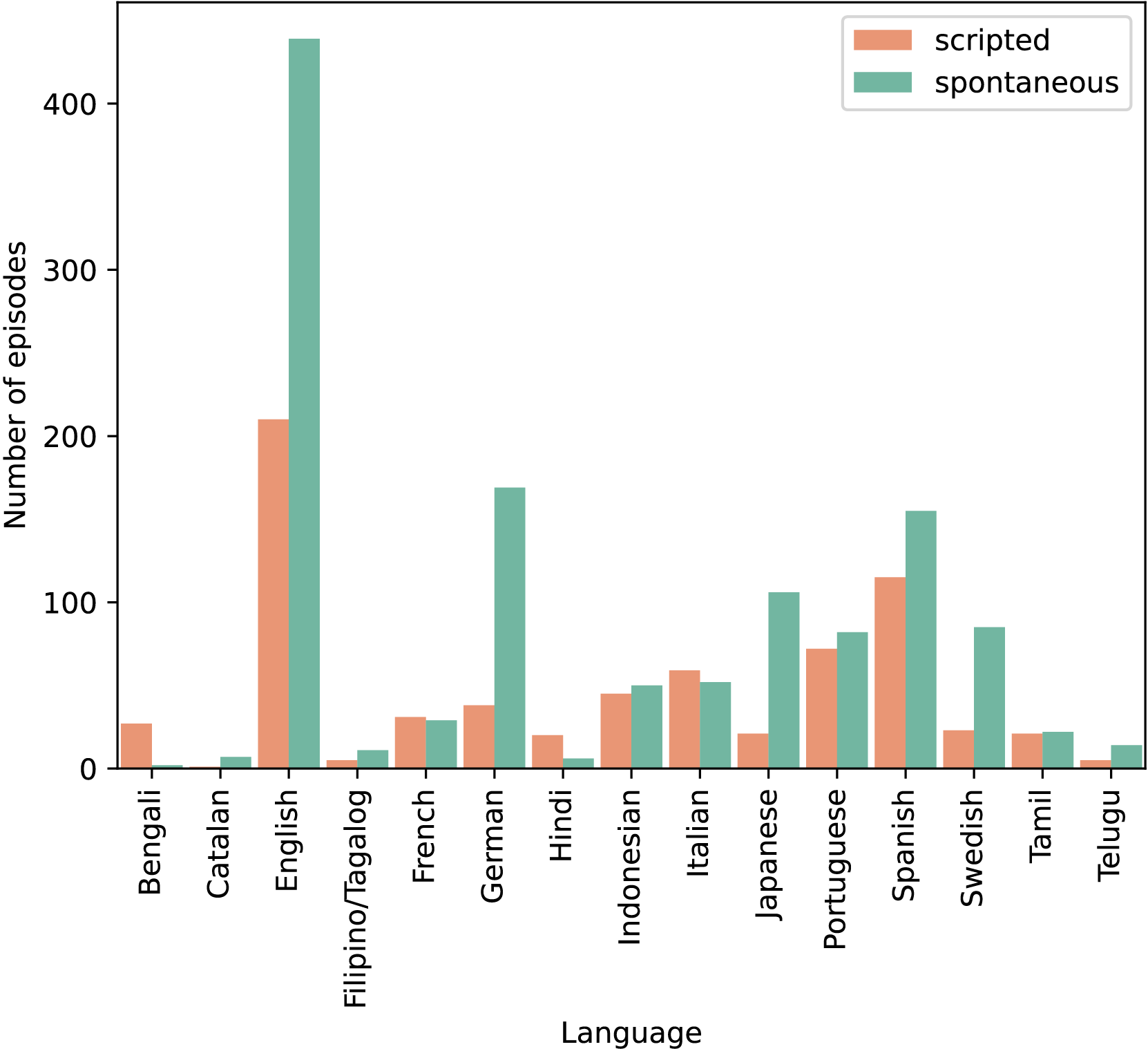}
  \caption{Distribution of spontaneous and scripted files in the podcast dataset split by language.}
  \label{fig:language_dist}
\end{figure}

\subsubsection{Podcast dataset}

We use a proprietary dataset of 4,000 of the most streamed Spotify podcast shows across 15 different markets, manually annotated by content experts for categories and formats. We assign ``\textit{Scripted}" and ``\textit{Spontaneous}" labels based on the format and category annotations. For example, we labelled shows annotated as ``Scripted narrative" or ``Scripted non-fiction" as ``\textit{Scripted}"; we labelled shows annotated as ``Blabbercast," ``Discussion," ``Improv" and ``Call-ins" as ``\textit{Spontaneous}." Shows with ambiguous or irrelevant formats (e.g. ``Interview", ``Soundscape") were filtered out. We include languages that have at least one show per label: \textit{Bengali, Catalan, English, Filipino/Tagalog, French, German, Hindi, Indonesian, Italian, Japanese, Portuguese, Spanish, Swedish, Tamil, and Telugu}. We further sample one episode per show (with the median audio duration). The final dataset contains nearly 700 \textit{scripted} podcasts and just over 1,230 \textit{spontaneous} podcasts. The language distribution is illustrated in Fig.~\ref{fig:language_dist}. The breakdown by format can be found in Table 1 in the Supplementary Materials (SM). We use this dataset to train and evaluate our models.

\subsubsection{Publicly available datasets}
\label{sec:external_data}
Due to the proprietary nature of this dataset, we also evaluate our models on two publicly available datasets:

\begin{enumerate}
    \item \textbf{Corpus d'Etude pour le Français Contemporain (CEFC)} \cite{cefc} is a dataset of approximately 450 hours of spoken audio in French by over 2,500 speakers. It combines data from 14 different sources, covering a range of communication styles and contexts (see SM:Table 1). The speakers come from various regions of France, Switzerland, and Belgium. This dataset is also used in \cite{christodoulides20_speechprosody}.
    \item \textbf{Third DIHARD Challenge Evaluation} \cite{dihard} is a dataset of approximately 33 hours of spoken audio in English and Mandarin Chinese. It contains political speeches, audiobooks, interviews, monologues, etc. The Mandarin Chinese sample is limited to ``web video," which we excluded from evaluation due to class ambiguity (SM:Table 1). This dataset is an evolution of the dataset used in \cite{ryant16_interspeech}.
\end{enumerate}

\noindent We assign the various speech styles a label of ``\textit{scripted}" or ``\textit{spontaneous}," as defined in SM:Table 1.

\subsection{Models}
\label{sec:features}

We extract a range of audio features and train neural network models to predict whether a podcast episode is scripted or spontaneous speech. We obtain a classification score on every 30 second snippet (the snippet duration used in Whisper~\cite{whisper}). Then, we apply an aggregation model to classify the episode as a whole.   

\subsubsection{Handcrafted features}

We extract handcrafted, acoustic features using the openSMILE tool~\cite{opensmile} and Pyannote~\cite{pyannote, pyannote2}, a speaker diarisation model. Using openSMILE, we extract the \textbf{\textit{eGeMAPSv02}} feature set on a functional level~\cite{opensmile_gemaps}: 88 acoustic features representing pitch, loudness, MFCCs, and duration of voiced and unvoiced segments, amongst other attributes, tailored for paralinguistic tasks (See Table 2 in SM). 

We supplement this default feature set with the mean and standard deviation of the speaking rate, extracted using the \textit{cPitchDirection} component in openSMILE. We also use Pyannote (\textit{pyannote/seg-mentation-3.0}) to extract summary statistics on the durations of predicted speech, non-speech, and overlapping speech segments (using pyannote's \textit{VoiceActivityDetection} and \textit{OverlappedSpeechDetection} pipelines)~\cite{pyannote, pyannote2}. We concatenate these additional features to the \textit{eGeMAPSv02} set for a \textbf{\textit{handcrafted}} feature set of dimension 115. Combined, these acoustic and prosodic features cover those evaluated in~\cite{christodoulides20_speechprosody, speaking_style_classification, ryant16_interspeech}. We standardise the feature set before feeding the input into the downstream architecture described in Section~\ref{sec:neural_net}.

\subsubsection{YAMNet}

\textbf{\textit{YAMNet}} is a deep learning model trained to predict 521 audio classes that cover a wide range of sounds, such as human vocals, musical instruments, animal and environmental sounds~\cite{howard2017mobilenets}. We extract class scores every 0.96s with a 50\% overlap, and process them using two separate methods:

\begin{enumerate}
    \item Statistical summaries: we calculate the mean and standard deviation of the scores per class across time, and concatenate the two vector features into a 1042 dimensional embedding.  
    \item mean-num-$4$: we count the number of frames a class was in the top-$4$ scores, as described in \cite{bbc_program}, to obtain a 512 dimensional vector feature.
\end{enumerate}

\subsubsection{Whisper}

We extract \textbf{\textit{Whisper}}~\cite{whisper} embeddings (specifically, \textit{whisper-large-v2}) by taking the last hidden state from the pre-trained model's encoder. This outputs a feature embedding of size 1,280 $\times$ 1,500 (i.e. a feature every 20ms of audio).

Additionally, we extract embeddings from a \textbf{\textit{fine-tuned (FT) Whisper}} model in the same manner as above. This model was fine-tuned on 800 hours of multilingual podcast audio ranging across 55 languages. There is no overlap between the podcast shows used for fine-tuning this model and the podcast dataset described in Section~\ref{sec:data}. This outputs a feature embedding of size 1,280 $\times$ 1,500.

\subsubsection{Neural network}
\label{sec:neural_net}

We segment each audio file in the podcast dataset into 30 second chunks, and run all the feature extraction models on each one. To avoid overfitting to abnormally long episodes, we sample exactly 25 sequential chunks (12.5 minutes) from the middle of each episode in the training dataset. We split our dataset into 5 folds on an episode level for cross-validation, stratifying by the podcast's category, format, and language to ensure equal representation in each fold. We train on the individual snippets by propagating the episode-level label down to the snippet level.

For each set of extracted features, we train 5 small neural networks: one per data split (4 folds as the training set and 1 fold for testing). The one dimensional feature sets (i.e. eGeMAPSv02, the handcrafted features, and the two YAMNet summaries) are each fed into a hidden layer of size 50 followed by a final binary classification layer. The two Whisper feature sets are two-dimensional, so are initially input into a hidden layer of size 100 and an average pooling layer, before connecting into the same architecture as the other feature models (inspired by the \textit{WhisperForAudioClassification} implementation in the transformers library\footnote{\url{https://github.com/huggingface/transformers}}). To deal with the class imbalance in our dataset, we initialise the bias in the final classification layer. The full architecture and training parameters are detailed in Table~\ref{tab:nn_architecture}. We save the best checkpoint per model based on validation loss for evaluation.

\begin{table}
    \centering
    \begin{tabular}{l|l}
 \textbf{Layer}&\textbf{Parameter}\\ \hline \hline
 Dense&100\\
 GlobalAveragePooling1D&\\ \hline
         Dense& 50\\
         Activation& ReLU\\
         Dropout& 0.2\\
         Dense& 1\\
         Activation& Sigmoid\\ \hline
         Adam Optimiser& 0.001\\
 Loss function&Binary Cross Entropy\\
 Metrics&Binary accuracy, AUC\\
 Epochs&Best of 40\textsuperscript{*}\\
 Batch size&64\\
    \end{tabular}
    \caption{Neural network architecture and parameters.~\textsuperscript{*}Whisper models were trained with a maximum of 10 epochs.}
    \label{tab:nn_architecture}
\end{table}
\subsubsection{Aggregation}

While we train the models on a snippet level, we add an aggregation step on the snippet-level predicted scores for a final episode-level score. We experiment with both median and mean aggregation functions.

\subsection{Evaluation}

We predict snippet scores on every chunk per episode and calculate a final episode-level score using the aggregation functions. Using the episode-level scores, we compute the average AUC score across the 5 folds, and the average F1-score per class with a classification threshold set to 0.5. We compare the different feature models to a majority-class (\textit{spontaneous}) baseline. We compute the same metrics on both the CEFC and DIHARD datasets (excluding examples we labelled as `ambiguous' in SM:Table 1) to understand how well the models perform cross-domain. Finally, we investigate language bias by computing the AUC per language. To allow for meaningful statistical analysis, we group low-resource languages into language-families (e.g.~\cite{goyal-etal-2020-efficient}); specifically, we group Bengali and Hindi into \textit{Indo-Aryan}, Telugu and Tamil into \textit{Dravidian}, and Filipino/Tagalog and Indonesian into \textit{Malayo-Polynesia}. We also remove Catalan from the language-specific analysis due to the small sample size.

\section{Results}
\label{sec:results}
\subsection{Overall performance}
\begin{table*}[t]

    \centering
    \begin{tabular}{l|c|c|c|c|c|c}  
         &    Majority &eGeMAPSv02&Handcrafted&  YAMNet&  Whisper& Fine-tuned Whisper\\  \hline
         Scripted F1-score&    0.00&0.69 $\pm$ 0.03&0.76 $\pm$ 0.03&  0.70 $\pm$ 0.02&  \textbf{0.83 $\pm$ 0.02}& 0.82 $\pm$ 0.03\\  
         Spontaneous F1-score&    0.78&0.86 $\pm$ 0.01&0.88 $\pm$ 0.02&  0.87 $\pm$ 0.01&  \textbf{0.91 $\pm$ 0.01}& \textbf{0.91 $\pm$ 0.01}\\  
         AUC&    0.50&0.87 $\pm$ 0.02&0.91 $\pm$ 0.02&  0.90 $\pm$ 0.01&  \textbf{0.95 $\pm$ 0.01} & \textbf{0.95 $\pm$ 0.01}\\ 
    \end{tabular}
    \caption{Overall AUC, Spontaneous F1-scores, and Scripted F1-scores of each feature model trained and evaluated on the podcast dataset. Scores are evaluated on an episode-level using median aggregation. Measures are an average and standard deviation across the 5-fold CV. Bold figures indicate the best performance per metric.}
    \label{tab:feature_results}
\end{table*}

\begin{figure}[t]
  \centering
  \includegraphics[width=\linewidth]{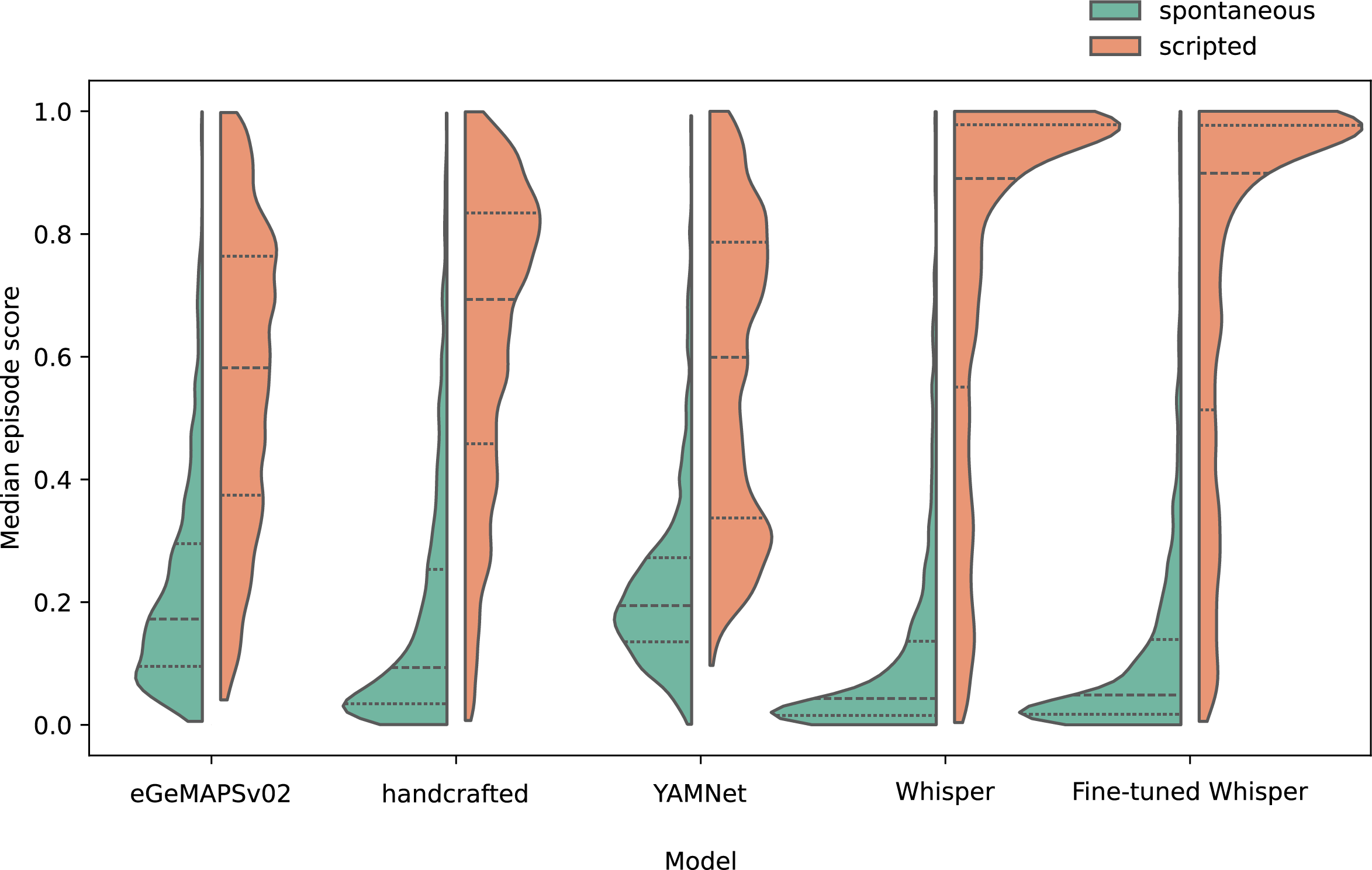}
  \caption{Distribution of median episode-level scores for scripted and spontaneous per feature model. Scores towards 1 are predicted as `Scripted' and towards 0 as `Spontaneous'.}
  \label{fig:score_dist}
\end{figure}

The overall performance on the podcast dataset is detailed in Table~\ref{tab:feature_results}. We find comparable performance between the two YAMNet preprocessing methods (i.e. statistical summaries and mean-num-$4$); we present the results for the statistical summaries model for brevity. Similarly, aggregating scores to an episode-level using mean versus median results in little to no difference - we report median aggregation. Whisper models are best at distinguishing scripted from spontaneous speech. This is in line with results across speech research (e.g.~\cite{latif2021sersurvey, ameer2023whisper, deepfake_detection_whisper, kodali23_interspeech, nonverbal_vocalisation}), where large transformer-based models show state-of-the-art performance compared to handcrafted, lower-level features. Given the large and diverse nature of the data Whisper was pre-trained on, this is unsurprising. Interestingly, fine-tuning Whisper does not yield better results over the pre-trained model. This suggests that podcasts, as a domain, are similar enough to the broader speech corpus used to train Whisper, at least in terms of this specific task. 

We note an improvement in performance when supplementing eGeMAPSv2 with additional features (Table~\ref{tab:feature_results}), specifically speech rate and summary statistics on the durations of predicted speech, non-speech, and overlapping speech segments. The difference is most pronounced in identifying scripted content, where the F1-scores increase from 0.69 to 0.76. This improvement aligns with findings from similar studies highlighting the saliency of speech and silence duration distributions and interactivity metrics, such as overlapping speech~\cite{speaking_style_classification, christodoulides20_speechprosody, ryant16_interspeech}. 

Lastly, we find that spontaneous F1-scores are higher than scripted F1-scores, regardless of the feature model (see Table~\ref{tab:feature_results}). This contrasts with the findings in~\cite{speaking_style_classification}, who reported higher accuracies on read speech classification. This is likely explained by the respective dataset distributions; our podcast dataset is imbalanced towards spontaneous speech, whereas the dataset analysed in~\cite{speaking_style_classification} is skewed towards scripted speech. For comparison, Whisper achieves average accuracies (=recall per class) of 95\% and 77\% on spontaneous and scripted speech, respectively, compared to the 69.5\% and 93.7\% reported in~\cite{speaking_style_classification}.

\subsection{Multilingual performance}

\begin{figure}[t]
  \centering
  \includegraphics[width=\linewidth]{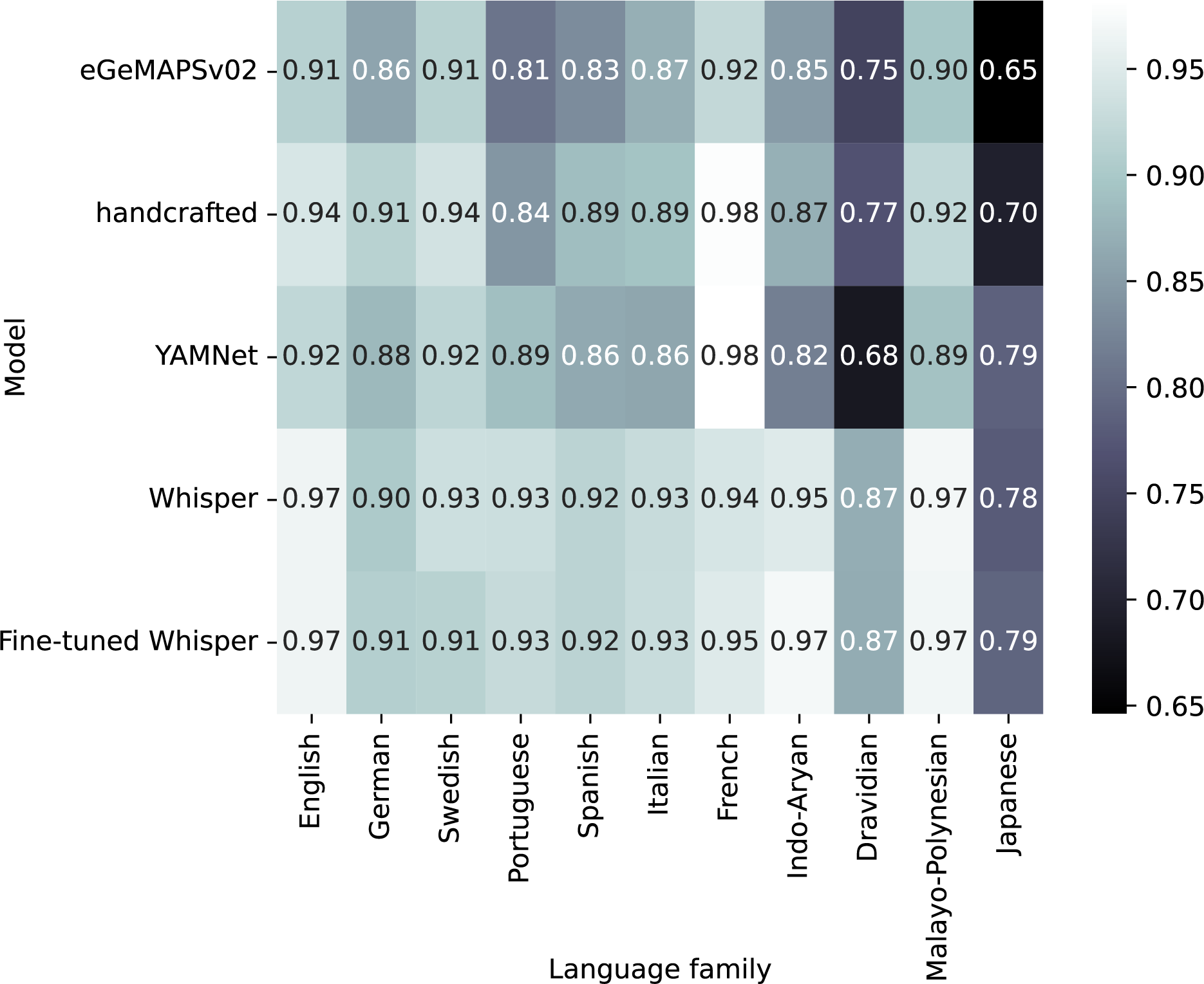}
  \caption{AUC computed on median episode-level scores for each feature model split by language.}
  \label{fig:auc_by_language}
\end{figure}

We break down the AUC performance for each method by language in Fig.~\ref{fig:auc_by_language}, where Whisper models tend to have better AUC per language. This is expected, as Whisper was exposed to 680,000 hours of speech, of which 117,000 hours cover a range of 96 non-English languages~\cite{whisper}. Unsurprisingly, English is one of the highest performing languages. All the models perform worse on Japanese speech; despite Whisper being pre-trained on 7,054 hours of Japanese speech (one of the higher-represented languages excluding English)~\cite{whisper}, it too struggles with Japanese. Most episodes have low score predictions (i.e. more ``spontaneous"), even those annotated as ``scripted". The training data distribution in Japanese is highly skewed towards spontaneous speech (see Fig.~\ref{fig:language_dist}), however both German and Swedish have similarly imbalanced sets and still perform near the overall AUC. The poor performance of acoustic and phonetic feature models (\textit{eGeMAPSv02} and \textit{handcrafted}) suggests either a discrepancy with the annotations, or a difference in salient features for distinguishing spontaneous from scripted speech in Japanese. Japanese falls under a different rhythm class of languages compared to the other languages in the podcast dataset; it is described as a ``mora-timed" language, compared to ``stress-timed" (e.g. English) and ``syllable-timed" (e.g. Spanish) languages. There is disagreement on the role rhythm plays in spontaneous Japanese speech: some studies report high variability in spontaneous speech between languages from different rhythmic classes (e.g. ~\cite{timing_differences_japanese}), whereas others reject the idea that mora use influences Japanese rhythm at all (e.g. ~\cite{japanese_mora}). Further investigation into the data is needed to understand the source of the errors; additional ``scripted" Japanese samples and evaluation on similar East-Asian languages can help clarify any discrepancies.

The Indo-Aryan and Malayo-Polynesian languages show comparable performance to English, despite belonging to different language families (see Fig.~\ref{fig:auc_by_language}). One possible reason is that these podcasts may actually include code-switching: where speakers alternate between multiple languages~\cite{code-switching}. \textit{Hinglish} (Hindi and English) and \textit{Taglish} (Tagalog and English) are common examples~\cite{code-switching_english}. Another potential explanation for the relatively high performance for Indo-Aryan (at least for the Whisper models) is that there is a much smaller sample and the data imbalance skews in the opposite direction (Fig.~\ref{fig:language_dist}): there are more scripted examples than spontaneous, the class that is supposedly easier to distinguish according to the literature~\cite{speaking_style_classification, christodoulides20_speechprosody, Tucker_Mukai_2023}. Analysing transcripts for the presence of English and evaluating the language performance by class can offer a more nuanced understanding.

\subsection{Cross-Domain Adaptability}
\begin{table}[t]
    \centering
    \begin{tabular}{l|c|c|c} 
         &  Podcast&  CEFC&  DIHARD\\ \hline
         eGeMAPSv02&  0.87 $\pm$ 0.02 &  0.53 $\pm$ 0.04 &  0.46 $\pm$ 0.06\\ 
         Handcrafted&  0.91 $\pm$ 0.02&  0.74 $\pm$ 0.03&  0.59 $\pm$ 0.02\\ 
         YAMNet&  0.90 $\pm$ 0.01 &  0.43 $\pm$ 0.01 &  0.55 $\pm$ 0.03\\ 
 Whisper& \textbf{0.95 $\pm$ 0.01}& \textbf{0.92 $\pm$ 0.02}&\textbf{0.95 $\pm$ 0.03}\\
 FT Whisper& \textbf{0.95 $\pm$ 0.01}& 0.91 $\pm$ 0.02 &0.92 $\pm$ 0.01\\
    \end{tabular}
    \caption{AUC computed on median file-level scores for each feature model across all datasets, evaluated using the labels assigned in SM:Table 1. Measures are an average and standard deviation across the 5-fold CV. Bold figures indicate the best performance per dataset.}
    \label{tab:cross_dataset}
\end{table}
    
We assess cross-domain generalisability by evaluating the models trained on the podcast dataset on other (non-podcast) speech datasets, namely CEFC and DIHARD (Section~\ref{sec:external_data}). While performance will vary depending on how the labels are assigned to the different speech sources, the AUC from each feature model is summarised in Table~\ref{tab:cross_dataset} for the label mappings defined in SM:Table 1. The file-level score distributions of all formats can be found in SM:Fig.1-3. Whisper generalises well across datasets, which is expected given the large and diverse dataset it was originally pre-trained on. On the other hand, eGeMAPSv02 and YAMNet perform poorly on the non-podcast datasets. YAMNet, in particular, struggles with classifying scripted content in DIHARD. While the supplemented features in the handcrafted feature set allow more generalisability over eGeMAPSv02 alone, performance is still much lower than the Whisper models. While these results could be explained by the model learning podcast-specific correlations, they may also be due to a difference in audio quality. This is because the podcast dataset is biased towards popular podcasts, which are typically created under professional recording conditions, whereas both CEFC and DIHARD have varied recording qualities. For example, meals and restaurant conversations may include background noise. This may also explain the slight decrease in performance when using the fine-tuned Whisper model compared to the pre-trained Whisper model.

\section{Conclusion}
\label{sec:conclusion}

We train a range of scripted and spontaneous speech audio-based classifiers on a large, multilingual podcast dataset. We compare the performance of handcrafted, interpretable features to modern, transformer-based speech representation models. Furthermore, we break down the performance by language to understand cross-lingual generalisability and biases. Finally, we assess the transferrability of a model trained on podcast recordings to other speech datasets. Our results show that transformer-based models, specifically Whisper, perform best at distinguishing speaking styles. More importantly, these models are better at generalising across languages and domains compared to handcrafted features and YAMNet. Despite the generally high performance across languages, certain languages, e.g. Japanese, are not modelled as well. While the source of the difference requires further investigation, it highlights the need to be aware of cultural differences when building large-scale models. One way to mitigate for such nuances is by conditioning the model based on language, for example by inputting the language as an additional feature. Another approach is to train language- or culture-specific models, at the expense of having several, less generalisable models. Analysis of such models can also offer additional insights into differences across languages. Models incorporating syntactic structure may also offer language-specific insights~\cite{spontaneous_syntax, pilan-etal-2024-conversational, syntactic_complexity_spontaneous}.

Transformer-based models show state-of-the-art performance across the speech processing research field; this experiment is no exception. However, the increased performance comes at the cost of interpretability and computational resources (e.g. the size of the handcrafted embedding compared to a Whisper embedding is 4KB to 7.3MB). Depending on the use-case, handcrafted features may be a good alternative for higher interpretability and understanding nuances (e.g. which acoustic features differ in the Japanese sample?).

Our results indicate that transformer-based models consistently outperform traditional feature-based techniques, achieving state-of-
the-art performance in distinguishing between scripted and
spontaneous speech across various languages. The models' ability to generalise beyond podcasts encourages further study into understanding speech style using this dataset. We plan to extend this work to capture more languages and cultures, encompassing a wider, more nuanced representation of speech styles and formats.  

\bibliographystyle{IEEEbib}
\bibliography{references}

\begin{thebibliography}{10}

\bibitem{spontaneous_or_read}
A.~Batliner, R.~Kompe, A.~Kie{\ss}ling, E.~N{\"o}th, and H.~Niemann,
\newblock ``Can you tell apart spontaneous and read speech if you just look at prosody?,''
\newblock in {\em Speech Recognition and Coding}, Antonio J.~Rubio Ayuso and Juan M.~L{\'o}pez Soler, Eds., Berlin, Heidelberg, 1995, pp. 321--324, Springer Berlin Heidelberg.

\bibitem{LAAN199743}
Gitta~P.M. Laan,
\newblock ``The contribution of intonation, segmental durations, and spectral features to the perception of a spontaneous and a read speaking style,''
\newblock {\em Speech Communication}, vol. 22, no. 1, pp. 43--65, 1997.

\bibitem{f0_speaking_style}
M.~Swerts, E.~Strangert, and M.~Heldner,
\newblock ``F/sub 0/ declination in read-aloud and spontaneous speech,''
\newblock in {\em Proceeding of Fourth International Conference on Spoken Language Processing. ICSLP '96}, 1996, vol.~3, pp. 1501--1504 vol.3.

\bibitem{Tucker_Mukai_2023}
Benjamin~V. Tucker and Yoichi Mukai,
\newblock {\em Spontaneous Speech},
\newblock Elements in Phonetics. Cambridge University Press, 2023.

\bibitem{martikainen2022exploring}
Katariina Martikainen, Jussi Karlgren, and Khiet~Phuong Truong,
\newblock ``Exploring audio-based stylistic variation in podcasts,''
\newblock in {\em Interspeech 2022}, 2022.

\bibitem{deldjoo2016content}
Yashar Deldjoo, Mehdi Elahi, Paolo Cremonesi, Franca Garzotto, Pietro Piazzolla, and Massimo Quadrana,
\newblock ``Content-based video recommendation system based on stylistic visual features,''
\newblock {\em Journal on Data Semantics}, vol. 5, pp. 99--113, 2016.

\bibitem{Deldjoo2020}
Yashar Deldjoo, Markus Schedl, Paolo Cremonesi, and Gabriella Pasi,
\newblock ``Recommender systems leveraging multimedia content,''
\newblock {\em ACM Comput. Surv.}, vol. 53, no. 5, sep 2020.

\bibitem{xie2022artistic}
Xin Xie, Yi~Li, Huaibo Huang, Haiyan Fu, Wanwan Wang, and Yanqing Guo,
\newblock ``Artistic style discovery with independent components,''
\newblock in {\em Proceedings of the IEEE/CVF Conference on Computer Vision and Pattern Recognition}, 2022, pp. 19870--19879.

\bibitem{WAGNER20151}
Petra Wagner, Jürgen Trouvain, and Frank Zimmerer,
\newblock ``In defense of stylistic diversity in speech research,''
\newblock {\em Journal of Phonetics}, vol. 48, pp. 1--12, 2015,
\newblock The Impact of Stylistic Diversity on Phonetic and Phonological Evidence and Modeling.

\bibitem{speech_variation_speaker_rec_implications}
Yoonjeong Lee and Jody Kreiman,
\newblock ``{Acoustic voice variation in spontaneous speech},''
\newblock {\em The Journal of the Acoustical Society of America}, vol. 151, no. 5, pp. 3462--3472, 05 2022.

\bibitem{audeering2021multimodal}
Andreas Triantafyllopoulos, Uwe Reichel, Shuo Liu, Stephan Huber, Florian Eyben, and Björn~W. Schuller,
\newblock ``Multistage linguistic conditioning of convolutional layers for speech emotion recognition,''
\newblock {\em Frontiers in Computer Science}, vol. 5, 2023.

\bibitem{luz20_interspeech}
Saturnino Luz, Fasih Haider, Sofia de~la Fuente, Davida Fromm, and Brian MacWhinney,
\newblock ``{Alzheimer’s Dementia Recognition Through Spontaneous Speech: The ADReSS Challenge},''
\newblock in {\em Proc. Interspeech 2020}, 2020, pp. 2172--2176.

\bibitem{williams19c_interspeech}
Jennifer Williams and Simon King,
\newblock ``{Disentangling Style Factors from Speaker Representations},''
\newblock in {\em Proc. Interspeech 2019}, 2019, pp. 3945--3949.

\bibitem{ethical_awareness_paralinguistics}
Anton Batliner, Michael Neumann, Felix Burkhardt, Alice Baird, Sarina Meyer, Ngoc~Thang Vu, and Björn~W. Schuller,
\newblock ``Ethical awareness in paralinguistics: A taxonomy of applications,''
\newblock {\em International Journal of Human–Computer Interaction}, vol. 39, no. 9, pp. 1904--1921, 2023.

\bibitem{speakerStyle_Laan_1997}
Gitta~P.M. Laan,
\newblock ``The contribution of intonation, segmental durations, and spectral features to the perception of a spontaneous and a read speaking style,''
\newblock {\em Speech Communication}, vol. 22, no. 1, pp. 43--65, 1997.

\bibitem{ryant16_interspeech}
Neville Ryant and Mark Liberman,
\newblock ``{Automatic Analysis of Phonetic Speech Style Dimensions},''
\newblock in {\em Proc. Interspeech 2016}, 2016, pp. 77--81.

\bibitem{christodoulides20_speechprosody}
George Christodoulides,
\newblock ``{Speaking Style Prosodic Variation and the Prosody-Syntax Interface: A Large-Scale Corpus Study},''
\newblock in {\em Proc. Speech Prosody 2020}, 2020, pp. 705--709.

\bibitem{speaking_style_classification}
Arlindo Veiga, Dirce Celorico, Jorge Proen{\c{c}}a, Sara Candeias, and Fernando Perdig{\~a}o,
\newblock ``Prosodic and phonetic features for speaking styles classification and detection,''
\newblock in {\em Advances in Speech and Language Technologies for Iberian Languages}, Berlin, Heidelberg, 2012, pp. 89--98, Springer Berlin Heidelberg.

\bibitem{opensmile_gemaps}
Florian Eyben, Klaus~R. Scherer, Björn~W. Schuller, Johan Sundberg, Elisabeth André, Carlos Busso, Laurence~Y. Devillers, Julien Epps, Petri Laukka, Shrikanth~S. Narayanan, and Khiet~P. Truong,
\newblock ``The geneva minimalistic acoustic parameter set (gemaps) for voice research and affective computing,''
\newblock {\em IEEE Transactions on Affective Computing}, vol. 7, no. 2, pp. 190--202, 2016.

\bibitem{schuller2018speech}
Bj{\"o}rn~W Schuller,
\newblock ``{Speech emotion recognition: Two decades in a nutshell, benchmarks, and ongoing trends},''
\newblock {\em Communications of the ACM}, vol. 61, no. 5, pp. 90--99, 2018.

\bibitem{whisper}
Alec Radford, Jong~Wook Kim, Tao Xu, Greg Brockman, Christine Mcleavey, and Ilya Sutskever,
\newblock ``Robust speech recognition via large-scale weak supervision,''
\newblock in {\em Proceedings of the 40th International Conference on Machine Learning}, Andreas Krause, Emma Brunskill, Kyunghyun Cho, Barbara Engelhardt, Sivan Sabato, and Jonathan Scarlett, Eds. 23--29 Jul 2023, vol. 202 of {\em Proceedings of Machine Learning Research}, pp. 28492--28518, PMLR.

\bibitem{wav2vec2}
Alexei Baevski, Henry Zhou, Abdelrahman Mohamed, and Michael Auli,
\newblock ``{wav2vec 2.0: A Framework for Self-Supervised Learning of Speech Representations},'' 2020.

\bibitem{latif2021sersurvey}
Siddique Latif, Rajib Rana, Sara Khalifa, Raja Jurdak, Junaid Qadir, and Bjoern~W Schuller,
\newblock ``Survey of deep representation learning for speech emotion recognition,''
\newblock {\em IEEE Transactions on Affective Computing}, pp. 1--1, 2021.

\bibitem{whisper_ser}
Erik Goron, Lena Asai, Elias Rut, and Martin Dinov,
\newblock ``Improving domain generalization in speech emotion recognition with whisper,''
\newblock in {\em ICASSP 2024 - 2024 IEEE International Conference on Acoustics, Speech and Signal Processing (ICASSP)}, 2024, pp. 11631--11635.

\bibitem{howard2017mobilenets}
Andrew~G Howard, Menglong Zhu, Bo~Chen, Dmitry Kalenichenko, Weijun Wang, Tobias Weyand, Marco Andreetto, and Hartwig Adam,
\newblock ``Mobilenets: Efficient convolutional neural networks for mobile vision applications,''
\newblock {\em arXiv preprint arXiv:1704.04861}, 2017.

\bibitem{bbc_program}
Lam Pham, Chris Baume, Qiuqiang Kong, Tassadaq Hussain, Wenwu Wang, and Mark Plumbley,
\newblock ``An audio-based deep learning framework for bbc television programme classification,''
\newblock in {\em 2021 29th European Signal Processing Conference (EUSIPCO)}, 2021, pp. 56--60.

\bibitem{ameer2023whisper}
Huma Ameer, Seemab Latif, Rabia Latif, and Sana Mukhtar,
\newblock ``Whisper in focus: Enhancing stuttered speech classification with encoder layer optimization,'' 2023.

\bibitem{deepfake_detection_whisper}
Piotr Kawa, Marcin Plata, Michał Czuba, Piotr Szymański, and Piotr Syga,
\newblock ``Improved deepfake detection using whisper features,'' 2023.

\bibitem{kodali23_interspeech}
Manila Kodali, Sudarsana~Reddy Kadiri, and Paavo Alku,
\newblock ``{Classification of Vocal Intensity Category from Speech using the Wav2vec2 and Whisper Embeddings},''
\newblock in {\em Proc. INTERSPEECH 2023}, 2023, pp. 4134--4138.

\bibitem{nonverbal_vocalisation}
Panagiotis Tzirakis, Alice Baird, Jeffrey Brooks, Christopher Gagne, Lauren Kim, Michael Opara, Christopher Gregory, Jacob Metrick, Garrett Boseck, Vineet Tiruvadi, Björn Schuller, Dacher Keltner, and Alan Cowen,
\newblock ``Large-scale nonverbal vocalization detection using transformers,''
\newblock in {\em ICASSP 2023 - 2023 IEEE International Conference on Acoustics, Speech and Signal Processing (ICASSP)}, 2023, pp. 1--5.

\bibitem{cefc}
ATILF, LIF, Loria, CLLE, ICAR, and Lattice,
\newblock ``Cefc,'' 2021,
\newblock {ORTOLANG} ({Open} {Resources} {and} {TOols} {for} {LANGuage}) \textendash www.ortolang.fr.

\bibitem{dihard}
Neville Ryant, Mark Liberman, James Fiumara, and Christopher Cieri,
\newblock ``Third dihard challenge evaluation,'' 2022.

\bibitem{opensmile}
Florian Eyben, Martin W\"{o}llmer, and Bj\"{o}rn Schuller,
\newblock ``Opensmile: The munich versatile and fast open-source audio feature extractor,''
\newblock in {\em Proceedings of the 18th ACM International Conference on Multimedia}, New York, NY, USA, 2010, MM '10, p. 1459–1462, Association for Computing Machinery.

\bibitem{pyannote}
Hervé Bredin,
\newblock ``{pyannote.audio 2.1 speaker diarization pipeline: principle, benchmark, and recipe},''
\newblock in {\em Proc. INTERSPEECH 2023}, 2023.

\bibitem{pyannote2}
Alexis Plaquet and Hervé Bredin,
\newblock ``{Powerset multi-class cross entropy loss for neural speaker diarization},''
\newblock in {\em Proc. INTERSPEECH 2023}, 2023.

\bibitem{goyal-etal-2020-efficient}
Vikrant Goyal, Sourav Kumar, and Dipti~Misra Sharma,
\newblock ``Efficient neural machine translation for low-resource languages via exploiting related languages,''
\newblock in {\em Proceedings of the 58th Annual Meeting of the Association for Computational Linguistics: Student Research Workshop}, Shruti Rijhwani, Jiangming Liu, Yizhong Wang, and Rotem Dror, Eds., Online, July 2020, pp. 162--168, Association for Computational Linguistics.

\bibitem{timing_differences_japanese}
Dan Brenner,
\newblock ``{Timing differences in read speech and spontaneous conversation: English, Japanese, Korean, and Mandarin},''
\newblock {\em The Journal of the Acoustical Society of America}, vol. 130, no. 4\_Supplement, pp. 2521--2521, 10 2011.

\bibitem{japanese_mora}
Natasha Warner and Takayuki Arai,
\newblock ``{The role of the mora in the timing of spontaneous Japanese speech},''
\newblock {\em The Journal of the Acoustical Society of America}, vol. 109, no. 3, pp. 1144--1156, 03 2001.

\bibitem{code-switching}
Peter Auer,
\newblock {\em Code-switching in conversation: Language, interaction and identity},
\newblock Routledge, 2013.

\bibitem{code-switching_english}
EDGAR~W. SCHNEIDER,
\newblock ``Hybrid englishes: An exploratory survey,''
\newblock {\em World Englishes}, vol. 35, no. 3, pp. 339--354, 2016.

\bibitem{spontaneous_syntax}
Jim Miller and Regina Weinert,
\newblock {\em {Spontaneous Spoken Language: Syntax and Discourse}},
\newblock Oxford University Press, 03 1998.

\bibitem{pilan-etal-2024-conversational}
Ildiko Pilan, Laurent Pr{\'e}vot, Hendrik Buschmeier, and Pierre Lison,
\newblock ``Conversational feedback in scripted versus spontaneous dialogues: A comparative analysis,''
\newblock in {\em Proceedings of the 25th Annual Meeting of the Special Interest Group on Discourse and Dialogue}, Tatsuya Kawahara, Vera Demberg, Stefan Ultes, Koji Inoue, Shikib Mehri, David Howcroft, and Kazunori Komatani, Eds., Kyoto, Japan, Sept. 2024, pp. 440--457, Association for Computational Linguistics.

\bibitem{syntactic_complexity_spontaneous}
Suma Bhat and Su-Youn Yoon,
\newblock ``Automatic assessment of syntactic complexity for spontaneous speech scoring,''
\newblock {\em Speech Communication}, vol. 67, pp. 42--57, 2015.

\end{thebibliography}

\end{document}